\documentclass{article}
\usepackage[margin=1.25in]{geometry}
\usepackage{titlesec}
\usepackage{parskip}
\usepackage{setspace}
\usepackage{graphicx}
\usepackage{makecell}
\date{}
\usepackage{xcolor}
\newcommand{\red}{\textcolor{black}}
\usepackage{booktabs}
\usepackage[notes,backend=bibtex]{biblatex-chicago}
\renewbibmacro*{cite:ibid}{\printtext[bibhyperref]{\bibstring[\mkibid]{ibidem}}}

\bibliography{calibration-final}

\usepackage{etoolbox}
\AtBeginEnvironment{quote}{\par\singlespacing\small}
\usepackage{arydshln}

\usepackage{amssymb} 
\usepackage{amsmath} 
\usepackage{bm}
\newtheorem{definition}{Definition}

\usepackage[utf8]{inputenc} 

\title{Does calibration mean what they say it means; or, the reference class problem rises again}
\author{Lily Hu\\forthcoming in \emph{Philosophical Studies}}
\titleformat{\section}
  {\normalfont\Large\bfseries}{\S\thesection}{1em}{}

\begin{document}
\maketitle
\begin{abstract}
Discussions of statistical criteria for fairness commonly convey the normative significance of \emph{calibration within groups} by invoking what risk scores ``mean." On the \emph{Same Meaning} picture, group-calibrated scores ``mean the same thing" (on average) across individuals from different groups and accordingly, guard against disparate treatment of individuals based on group membership. My contention is that calibration guarantees no such thing. Since concrete actual people belong to many groups, calibration cannot ensure the kind of consistent score interpretation that the Same Meaning picture implies matters for fairness, unless calibration is met within every group to which an individual belongs. Alas only perfect predictors may meet this bar. The Same Meaning picture thus commits a \emph{reference class fallacy} by inferring from calibration within some group to the “meaning” or evidential value of an individual’s score, because they are a member of that group. The reference class answer it presumes \red{does not only lack justification; it is very likely wrong}. I then show that the reference class problem besets not just calibration but other group statistical criteria that claim a close connection to fairness. Reflecting on the origins of this oversight opens a wider lens onto the predominant methodology in algorithmic fairness based on stylized cases.
\end{abstract}

\section{Introduction}
In the literature on algorithmic fairness, group-based statistical parity conditions take stock of how an algorithm’s predictions square with true outcomes for each subpopulation or group of interest. Picking-and-choosing, mixing-and-matching different error quantities and outcome- or prediction-defined populations across different groupings yields a slew of statistical measures, parity across which has conventionally been suggestive of a kind of equal, and thus fair, treatment and lack of parity indicating divergent, possibly unfair, treatment. Within this multidisciplinary discussion, philosophers have taken up the mantle of asking which, if any, of the statistical criteria that have been proposed in technical contributions are genuinely closely tied to fairness. Some authors deflate the significance of certain statistical conditions in favor of alternatives;\autocite{long2021fairness,hedden2021statistical,loi2022calibration} while others aim to shore up the status of those criteria diminished by their fellow philosophers.\autocite{hellman2020measuring,grant2023equalized} Some contributions strive to reconcile existing fairness criteria which seem to conflict;\autocite{beigang2023reconciling,holm2023fairness} while their less optimistic peers lament the slim pickings and submit criteria of their own for consideration.\autocite{eva2022algorithmic} By and large, findings within this literature have been broadly negative: of the proposed conditions, few, it seems, are possibly necessary, and none sufficient for fairness. 

Still, among the candidate criteria, \emph{calibration within groups} has retained a normative edge, holding on most strongly to a claim to being a true requirement for fairness. Calibration speaks only to facts about model behavior, whereas other measures are sensitive to factors presumably irrelevant to fairness such as the shape of a population’s risk distribution. While other metrics are defined by reference to particular uses of an algorithm to make decisions, calibration assesses a tool’s overall performance, capturing whether across all of its possible outputs, the model is equally well-fit (on average) to the target outcome for different subpopulations or groups (e.g., by race, sex, age). Calibration within groups is thus thought to secure for decision-makers a kind of sound and consistent interpretation of risk scores, such that they are, as Kleinberg et al. put it in their landmark paper, “justified in treating people with the same score comparably with respect to the outcome.”\autocite{kleinberg2016inherent}

This property of calibration within groups is often glossed in the literature as ensuring that scores “mean the same thing” for individuals of different groups. Brian Hedden, for instance, writes that risk scores that are calibrated within groups, “mean the same thing, or [] have the same evidential value, regardless of the group to which the individual belongs.”\autocite[216]{hedden2021statistical} Scores that are miscalibrated, on the other hand, “mean different things” for different groups, leaving decision-makers at risk of weighing the costs of error differently across similar individuals, thereby potentially “treating individuals differently in virtue of their group membership."\autocite[225]{hedden2021statistical}  

My aim in this paper is to draw out this widespread interpretation of calibration within groups, which I call the \emph{Same Meaning} picture, and show why its portrayal of calibration's distinctive normative value and its comparative advantage over other group-based statistical criteria is false. Its central claim is that because risk scores that are calibrated within groups mean the same thing (on average) across individuals from different groups, they allow for a kind of sound and even interpretation of individuals' scores and thereby guard against decision policies, which engage in systematic disparate treatment based on the group to which an individual belongs. My contention is that calibration, unfortunately, can guarantee no such thing—or at least, it cannot guarantee the version of the Same Meaning view that is normatively significant for the fairness of an algorithmic system. Arguments that it does commit a critical error: since concrete actual people belong to many groups, calibration cannot ensure the kind of consistent interpretation of scores that the Same Meaning picture implies matters for fairness—unless calibration is met within every group to which an individual belongs. Alas only perfect predictors meet this bar. The Same Meaning motivation for calibration thus commits a \emph{reference class fallacy} by presuming that calibration within some pre-specified group speaks to the “meaning” or evidential value of a particular individual’s score vis-à-vis their ``true" probability, simply because they may be classified as a member of that group. This is precisely the problem of reference classes: \emph{Which} of the many group probability facts corresponding to the many groups to which an individual belongs should apply when figuring the probability of their individual case? \red{Thus, the Same Meaning picture claims an answer to the reference class problem, but it provides no argument for that answer.} What is more, the reference class answer that the Same Meaning picture assumes is, in all likelihood, wrong. Or so I will argue.

Showing ``meaning"-based talk about calibration to rely on a mistake undercuts a premise that undergirds much debate across disciplines about the property's normative significance. For the {Same Meaning} picture of calibration is invoked not just by its proponents who take it to have distinctive normative import compared with competitor criteria but also by calibration’s detractors. It is propagated not only by philosophers and legal scholars looking to convey its relevance for fairness; it is also expressed in foundational technical texts in algorithmic fairness. 

But it is not this paper’s primary concern to debunk calibration within groups. The story about calibration serves as an entry point into a broader investigation of the debate on statistical criteria of fairness. Identifying the mistake within commentary about calibration draws the curtain to reveal a fractal pattern, where the reference class problem is peppered throughout discussions of fairness, rearing its head in other motivations given for the normative significance of group statistical facts. As I will show, it afflicts all attempts to connect group-based statistical criteria with a notion of fairness that speaks to individuals. Furthermore, reflecting on the origins of this error opens a wider lens onto algorithmic fairness, identifying risks and pitfalls in the predominant methodology based on stylized cases that authors have taken in theorizing statistical criteria of fairness.

I proceed as follows. I start in \S 2 by formally introducing calibration within groups, characterizing the Same Meaning picture, and presenting my challenge to it based in the problem of reference classes. In \S 3, I move to clarify and then undercut the presumptive connection between calibration's ``same meaning" property and fairness in algorithmic systems. This will in turn address two replies that seek to sidestep the reference class charge by either modifying the Same Meaning picture or altering the notion of fairness that calibration allegedly safeguards. In \S 4, I set my sights beyond calibration. A prominent critique of the condition by Benjamin Eva serves as a point of departure for a broader challenge to prevailing approaches to statistical criteria for fairness. I close with some remarks on the hazards of excess abstraction in theorizing algorithmic fairness and conclude in \S 5.

\section{Calibration and ``meaning"-based talk}
Discussions of calibration within groups as a candidate fairness criterion typically begin by introducing a more general notion of calibration, long taken by practitioners to be a desideratum of adequate risk scoring systems. A system that is well-calibrated on this account will generate risk scores that track actually observed risk or the true probability or relative frequency of the target outcome on average. That is, for the set of individuals who receive a given risk score $r$, the expected proportion of them that has (or will have) the outcome $Y$ is equal to $r$.   

\begin{definition}[Calibration]

An individual $X \in \bm{X}$ receives a risk score r according to a scoring function $s(X): X \to [0, 1].$ Risk scores are \emph{calibrated} if $\forall r \in [0, 1]$,
\[ P(Y(X) = 1 \mid s(X) = r ) = r.\]
\end{definition}

This classical notion of calibration is defined over the entire population of individuals scored by the algorithm. But calibration may also be assessed with respect to subpopulations to ask whether, for a given subset of individuals scored by the algorithm, the model’s outputs track \emph{their} true probabilities, on average, of the outcome of interest. Scores calibrated for one grouping and miscalibrated for another suggest that a model's predictions are better fit to the former than the latter, and this uneven performance on different groups has been thought by contributors to the literature to be a mark against an algorithm’s fairness. This idea motivates a fairness-related notion of calibration commonly termed in the literature, “calibration within groups.”

\begin{definition}[Calibration Within Groups]\footnote{A weaker notion of calibration within groups does not require that scores calibrated within groups also track true probabilities or actual relative frequencies in the group, \red{thus dropping the final ``$=r$'' in Definition 2:
\[P(Y(X) = 1 \mid s(X) = r, d(X) = d_1 ) = P(Y(X) = 1 \mid s(X) = r, d(X) = d_2).\]}
My argument against the Same Meaning view applies to both formulations of calibration within groups, so opting for the definition here makes no difference to the bulk of what I say in this paper.}

An individual $X \in \bm{X}$ belongs to a group $d \in D$ $\red{\subset P(\bm{X})}$  and receives a risk score $r$ according to a scoring function $s(X): X \to [0, 1]$. Risk scores are \emph{calibrated within $\bm{D}$} if $\forall r \in [0,1]$ and any $d_1, d_2 \in D$,
\[P(Y(X) = 1 \mid s(X) = r, d(X) = d_1 ) = P(Y(X) = 1 \mid s(X) = r, d(X) = d_2) = r.\]
\end{definition}
While calibration \emph{neat} (Definition 1) may be satisfied by an imperfect predictor, only perfect predictors may satisfy calibration within groups for all groups. To see this point, notice that for any imperfect predictor, one can define a group that contains only those individuals on whom the predictor makes errors. Scores will fail to be calibrated across that group and its complement. Since it is this second kind of calibration notion that is suggested in the literature to have distinctive moral relevance for the fairness of a risk scoring system, for the remainder of the paper I’ll for the most part use “calibration” as shorthand for “calibration within groups.” Where I mean to refer to the classical notion, I will make clear that I am discussing calibration \`{a} la Definition 1. 
\subsection{The Same Meaning picture}

Risk scores that are calibrated within groups ensure that a model is, on average, equally well-fit to those groups vis-à-vis the outcome that it predicts. This property is often glossed in the literature as ensuring that scores “mean the same thing” for individuals of different groups.

The Same Meaning picture is intuitive to grasp, but it is worth lingering to note two features of its portrayal of calibration's connection to fairness. First, the invocation of what scores ``mean" positions the reader in the seat of \emph{score interpretation}, highlighting its significance as a site of normative concern. Risk scores should dictate in a straightforward way what, in Robert Long's words, “rational credence” a decision-maker should have “that an individual has the unknown property.”\autocite[62]{long2021fairness} It should be clear what a score indicates about an individual's likelihood of bearing the outcome at hand. Decision-makers tasked with interpreting an algorithmically-output score should easily be able to answer the question, posed by Deborah Hellman, ``Given this evidence... what should I believe?"\autocite[828]{hellman2020measuring}

The hope is not just to be able to answer this question but to do so \emph{consistently} across the many scores a decision-maker comes to face with. What we want, as Brian Hedden writes, is for ``assignment of a given risk score to have the same evidential value, regardless of the group to which the individual belongs.”\autocite[216]{hedden2021statistical} Scores that are calibrated carry the ``same meaning" and therefore, as Sandra Mayson writes, “communicate the same average risk regardless of the [group] of the person to whom it applies.”\autocite[2218]{mayson2019bias} The Same Meaning picture pumps the intuition for why despite its being a group-based parity metric, calibration compared against competitor criteria seems able to secure a distinctive kind of fair treatment for {individuals}. For it does not just generically contribute to the total evidence one has about a tool's behavior and an individual's likely outcome, as other group-based statistical criteria do. Rather, it suggests a more specific guarantee to the parties involved in risk assessment: that the judged receive scores that are true to \emph{them} and not improperly distorted by their group membership and \red{so the judge may take scores at face value and compare different scores with ease.}

By contrast, one who must contend with miscalibrated scores, which carry ``different meanings," is apprehensive. Hellman captures this anxiety well: “If a high-risk score means something different for blacks than for whites, then we do not know whether to believe (or how much confidence to have) in the claim that a particular scored individual is likely to commit a crime in the future.”\autocite[17]{hellman2020measuring} Unease about proper score interpretation ramifies into worries about fair treatment---when ``scores [] mean different things for different racial groups," Josh Simons writes, ``it is hard to see how they can treat people as equals when using [the tool] to make decisions”\autocite[43]{simons2023algorithms}---and distortionary effects on decision policies. So, Pleiss et al. in a machine learning article titled ``Fairness and Calibration" worry that a tool that is ``not calibrated with respect to groups defined by race... [may] have the unintended and highly undesirable consequence of incentivizing judges to take race into account when interpreting its predictions" and ``be misleading in the sense that the end user of these estimates has an incentive to mistrust (and therefore potentially misuse) them."\autocite[1, 2]{pleiss2017fairness}

``Meaning"-based glosses on calibration comport with the intuitive thought that a fair risk scoring algorithm is one that treats individuals who are similar with respect to the outcome similarly. Decision policies based on calibrated scores ensure that all individuals with a given score are treated according to their true probability of facing the outcome at issue, regardless of group membership. Scores that are miscalibrated, meanwhile, facilitate systematically biased inferences about risk and in turn support policies that are “equivalent to implementing inappropriate decision thresholds, which subject[] individuals to inappropriate disparate treatment.”\autocite[62]{long2021fairness}

I will now argue that calibration within groups cannot ensure the kind of consistent interpretation of scores that is portrayed in the Same Meaning picture, for the simple reason that actual individuals scored by an algorithm belong to \emph{many} groups. This fact appreciably complicates what calibration (and miscalibration) means for what a score-interpreter should believe about \red{an individual’s true risk given their risk score} and accordingly, the extent to which calibration may secure the particular kind of fairness evoked in the Same Meaning picture.

\subsection{The Same Meaning picture’s reference class problem}
The “same meaning” gloss on calibration is often presented from the perspective of an agent who is faced with an algorithmically-output risk score, tasked with interpretation and decision. If the scores she receives are calibrated within groups, the story goes, she may rest easy (at least as pertains to this source of potential unfairness); if they are miscalibrated, she has cause for worry. It is illuminating to hear Hedden on this latter situation at length. Calibration, he writes, 
\begin{quote}
is intuitively compelling and easily motivated. If it were violated by some algorithm, that would mean that the same risk score would have different evidential import for the two groups. Our probability that an individual is positive, given that they received a given risk score, would have to be different depending on the group to which the individual belongs. A given risk score, intended to be interpreted probabilistically, would in fact correspond to a different probability of being positive, depending on the individual’s group membership. This seems to amount to treating individuals differently in virtue of their differing group membership\autocite[225--226]{hedden2021statistical}.
\end{quote}

Like other invocations of the Same Meaning picture, the passage above starts from a \red{group-based probabilistic fact to make a claim about the rational credence that one should have in individual cases. The claim is that the fact that the algorithmically-output risk scores are (mis)calibrated within some group (e.g., race) speaks to the “meaning” or evidential value of particular individuals’ scores. The problem is that proponents of the Same Meaning picture provide no argument to back this inference from group probabilistic fact to individual probability. They simply \emph{assume} that the groups within which calibration is satisfied are indeed those groups that grant this inference.}

It is important to stress from the outset that the error I am pointing out in Hedden’s explication of calibration is \emph{not} about mistakenly taking group ratios or group-based probabilities to directly carry over and apply to individuals in that group. It is well-noted in the algorithmic fairness literature that group-based classification parity measures do not directly characterize probabilities for individuals who belong to the group.\footnote{The point that equalizing population-level ratios does not equalize individual-level chances is made forcefully in \autocite{castro2023correction}.} (Hence why practitioners often note that classic calibration of a predictive tool is a weak guarantee of predictive accuracy.) What is more, that problem could be resolved by taking care to note that calibration within groups ensures that individual scores “on average” or “in expectation” mean the same thing, regardless of group membership. 

But that recasting does not solve \emph{my} problem, for it still presupposes that calibration within that \emph{pre-specified group} speaks to the “meaning” or evidential value of a \emph{particular individual’s} score---\red{and it is this inferential leap that is my target. Without further defense, it is unjustified and, indeed, an instance of a familiar fallacy in probabilistic reasoning}: the assumption that a probabilistic fact about a group is relevant to the corresponding probability for some individual, for the sole reason that the individual may be classified as a member of the group. This is, of course, that venerable bugbear of probability theory, the \emph{reference class problem}, a problem that traces back to John Venn and Hans Reichenbach (who coined the problem), and has troubled statisticians, decision theorists, and ordinary people attempting rational inference and action alike ever since.\autocite{venn1888logic,reichenbach1971theory}

We face a reference class problem whenever we want to assign a probability to a particular individual case, but the case at hand may be classified in many different ways, each of which suggests for it a different probability. So, we must ask: which of these different conditional probabilities referring to different reference classes gives the right unconditional probability? It is a problem that, as argued in an influential paper by Alan H\'{a}jek, besets all normative theories of probability and so afflicts whatever interpretation of probability that one takes of algorithmically-generated risk scores meant to guide decision-making.\red{\footnote{\autocite{hajek2007reference}. \red{Most authors who write on algorithmic fairness do not defend a particular account of the probabilities encoded in statistical fairness criteria. Those who do make mention of different interpretations of probability are ecumenical. Hedden, for instance, takes them in the vein of objective chances in contexts where objective chanciness is involved and ``epistemic probabilities, and more specifically as the subjective probabilities that would be assigned by a reasonable individual who is familiar with the workings of the algorithm in question" in other contexts. ``On statistical criteria,"  214. Eva follows Hedden and writes: ``Generally, I think of these expectation values as being computed relative to objective probability functions \red{that encode either physical chances, long run frequencies, or the subjective probabilities of some suitably idealized observer.'' ``Algorithmic fairness and base rate tracking," 242.} I take a similarly ecumenical approach to probabilities in this paper.}}} The problem has special bite in cases where questions of algorithmic fairness arise, since except in cases of perfect prediction, individuals will belong to both groupings within which scores are calibrated and groupings within which scores are miscalibrated. And while a recent flurry of technical work in ``multicalibration" seeks to improve upon the relatively weak guarantees given by single-grouping-based calibration by ensuring that scores are calibrated in a collection of subgroupings (including intersections of multiple coarser groupings), these works make progress on the matter only insofar as they approach a solution to the reference class problem.\footnote{So long as an algorithm is predictively imperfect, errors will be distributed in some way across various groupings of individuals, leaving some still miscalibrated and so the reference class problem still a problem.} The problem for the Same Meaning picture is that interpreting the evidential value of a given individual’s score and comparing individuals' scores requires figuring \emph{which} groupings and accordingly \emph{which} calibration facts should apply for each. Increasing the number of reference classes accounted for in calibration within groups may improve an algorithm's predictive accuracy. But without claiming to have solved the reference class problem for all individuals scored by the algorithm, calibration cannot ensure that scores mean the same thing.\footnote{I should flag here two technical contributions that note a connection between calibration and the reference class problem. \autocite{roth2023reconciling}; \autocite{holtgen2023richness}. Both works see multicalibration as primarily fruitful for improving a model's predictive performance and solving  technical problems in the literature. I claim, contra these authors, that calibration's normative edge derives from its claim to sound and consistent score {interpretation} and comparison, a thicker picture of its relevance for fairness than one that speaks only to improved accuracy. Multicalibration may increase the likelihood that individual predictions are correct, but this does not ensure that scores ``mean the same thing"—sans a solution to the reference class problem.} 




It is curious that the reference class problem, a problem rather well-known to statisticians and epistemologists alike, rears its head with little notice here and, as I will show in §4, elsewhere in the philosophical literature on statistical criteria for fairness. In the case of the Same Meaning picture, the problem lies hidden in large part, I think, due to the highly abstracted vignettes that conventionally feature in “meaning”-based explications of calibration. In most commentary on calibration and fairness, scores are appended to individuals depicted as having only a single feature or descriptor that applies to them. Individuals are only racialized or sexed; they are just “Black” or “white,” or they are “male” or “female.” In toy cases, they belong to one of two rooms (Hedden), or to one of two course sections (Long). And so on. When problems are formulated as having only one variable of interest, the question, “To what group does this individual belong?” does not, indeed \emph{cannot}, arise. In a world with only one variable, calibration within that variable appears to speak to what it is rational to believe about that individual characterized by their race, sex, room, course section. The reference class problem does not appear, because the problem setup does not afford it.

Meanwhile, real concrete persons scored by algorithms are not so thinly characterized. No actually existing person has only a racial status or only a sex status or only an age. And rarely are those tasked with interpreting risk scores presented with just a single fact about the scored individual. (Though even were this so, it would only mean that the limited information available forces agents to “resolve” the reference class problem a certain way; it speaks nothing of whether this is a good way of doing so and so does little to protect against the kind of unfairness that calibration supposedly steers an algorithm and score-interpreter clear of.) When an individual belongs to many groups—when they are not only Black but also male and also 35 years-old and so also a 35 year-old Black male—\red{it becomes clear that calibration within groups can only speak to what their score means when it picks out the right group for the individual: that is, the right reference class.}

\red{To see more clearly why calibration cannot tell us what scores mean without an answer to the reference class problem, consider a case where} an algorithm is calibrated within racial and age groupings but not within sex groupings. How should one make sense of the same meaning claim in this case? That \emph{qua} a racialized person and \emph{qua} a person with an age, Jamal’s score of 8 “means the same thing” (in expectation) as other 8s of non-Black, non-35-year-olds? But that as the score of a male-sexed individual, his 8 indicates a different probability (in expectation) compared with those who are female-sexed with a score of 8? Does the score of 8 assigned to Jamal, a 35 year-old Black male, on average “mean the same thing” as a score of 8 assigned to Emily, a 20 year-old white female? Calibration's meaning-based talk confusingly suggests at once, “yes” and “no”: yes, as members of different racial groups and age groups; no, as members of different sex groups. 

\red{Nor does calibration help us interpret what individual scores mean non-comparatively.} It will here pay to build the case out in a way popular among contributors to the algorithmic fairness literature, by associating group probabilities (or ratios) with various groupings of scored individuals. Consider Emily, an individual who belongs to the group represented by the last row in the table below.\red{\footnote{\red{I use this example as an illustration of an algorithm that meets calibration within some groups but fails to make good on the Same Meaning picture’s articulation of the normative significance of this criterion. I make no claims about whether the algorithm is, on the whole, fair or accurate or legally permissible or otherwise a ``good" predictor.}}}

  \begin{center}
  \begin{tabular}{ cccc||c }
  Race & Age & Sex & Risk Score & Group Probability  \\ \Xhline{1\arrayrulewidth}
    Black & 35 & Male & 8 &  0.8 \\  
  Black & 35 & Female & 8 & 0.8  \\   \hdashline[1pt/2pt]
    Black & 20 & Male & 8 &  1 \\ 
  Black & 20 & Female & 8 & 0.6 \\\noalign{\smallskip} \hdashline \noalign{\smallskip}
    White & 35 & Male & 8 & 1 \\  
  White & 35 & Female & 8 & 0.6 \\   \hdashline[1pt/2pt]
    White & 20 & Male & 8 & 0.9  \\
  White & 20 & Female & 8 & 0.7 \\ 
\end{tabular}\end{center}

On the simplifying assumption that each group represented by a row contains equal numbers of people, this algorithm is calibrated in the way described earlier: within racial and age groupings (indeed, it is calibrated within ``intersectional" racial and age groups) but not within sex groupings. \red{The group of Black individuals assigned a score of 8 has an 80\% chance of the outcome; the same is true of the group of white individuals assigned a score of 8. The group of 35 year-old individuals with a score of 8 also has an 80\% chance of the outcome; the same is true for the group of 20 year-old individuals assigned an 8. On the other hand, the tool's violation of calibration within sex groupings is evinced by the fact that the group of those sexed male with a score of 8 has a 92.5\% chance of the outcome, but the group of those sexed female with a score of 8 has only a 67.5\% chance.}

Calibration supposedly speaks to what scores mean, but what \emph{does} Emily's score of 8 mean? \red{In other words, given that her risk score is 8, what we should take her individual probability to be? If we take Emily as a white person with a score of 8, we may rationally infer that she has an 80\% chance of the outcome. Taking her as a 20 year-old with a score of 8 also suggests an 80\% chance. But if we take Emily as someone sexed female assigned an 8, we would rationally infer that she has a 67.5\% chance. And if we take her as someone who is white, female, 20 years-old, and assigned a score of 8, then we would take her individual probability to be 70\%. The question then is, which grouping, each of which contains Emily as a member, carries the correct probability for her case? This is just the reference class problem. Thus, calibration makes headway on what Emily's score of 8 means for her individual probability only insofar as it has an answer to the reference class problem.}\footnote{The following table brings out the algorithm's violation of calibration within sex groupings.

  \begin{center}
  \begin{tabular}{ c c c c || c }
  Sex &Race & Age &  Risk Score & Group Probability \\ \Xhline{2\arrayrulewidth}
   Male & Black & 35 & 8 & 0.8  \\  
  Male & Black & 20 & 8 & 1  \\   
    Male & White & 35 & 8 & 1  \\ 
  Male & White & 20 & 8 &  0.9 \\\noalign{\smallskip} \hdashline \noalign{\smallskip}
    Female &Black & 35 & 8 &  0.8  \\  
  Female & Black & 20 & 8 & 0.6  \\  
   Female & White & 35 & 8 &  0.6 \\
  Female & White & 20 & 8 & 0.7 \\ 
\end{tabular}\end{center}}


In sum, the Same Meaning picture of calibration’s distinctive normative edge hinges on a solution to the reference class problem, a solution that is presupposed and never explicitly defended. The inference that calibration within groups ensures that individual scores “mean the same thing” as other scores is only safely drawn on an antecedent determination that \emph{that grouping} within which scores are calibrated is the right reference class for the scores about which one is reasoning. In encouraging us to conclude as much, the Same Meaning picture claims an answer to the reference class problem \red{but nowhere presents an argument for this answer}.\footnote{The inference that miscalibrated scores ``mean different things" is similarly subject to a reference class fallacy charge when authors draw the inference via statistical facts about two groups to which the individuals belong. What is true, however, is that if scores are miscalibrated within groups, then there exists some pair of individuals whose scores ``mean different things" (in expectation). But this follows from classic calibration's violation, not from miscalibration within groups in particular.}



Notice that even while the reference class problem besets all attempts at inference and decision-making based on probabilities and so is, as they say, ``everyone's problem," the Same Meaning picture engages it in a unique way. For it does not simply assume that we can solve or do our best to solve the problem of reference classes but to have \emph{in fact} solved it. That is, the Same Meaning picture of calibration is in effect a claim to a \emph{particular} solution. \red{For before, so the story goes, we did not know how to interpret individuals' risk scores; we were concerned that an 8 for Jamal might mean something different for his individual probability than the same score would mean for Emily's individual probability.} But the proponent of calibration's Same Meaning picture tells us not to worry. The algorithm is calibrated within (racial) groups, and so we may trust that the two scores ``mean the same thing." This is just a claim to have solved the reference class problem.  

An incredible feat were it so. I want to now move to underscore how incredible such a solution would be. For its claim would be not just to have solved the reference class problem for a particular individual case—not just to have solved it for, say, Emily and so determined the evidential value of Emily's score—but to have solved it for \emph{all individuals' scores}, thereby allowing scores to be compared with each other. That is to say that the Same Meaning picture assumes that the groups within which scores are calibrated encompass the right reference class for every individual, each of which is likely to be different from the next. Following a standard thought in the reference class literature that traces back to Reichenbach himself, this would mean that no features unaccounted for in the groups within which scores are calibrated supply statistically relevant information for \emph{any} of the individuals being scored.\footnote{Still, this proposed ``solution" to the reference class problem (select the narrowest class that contains all relevant information and nothing extraneous) faces serious obstacles and at best offers a partial heuristic. See \autocite{reichenbach1971theory,salmon1971statistical,thorn2017preference}.} 

An example serves to bring out the implausibility of such a claim. If calibration within racial groups ensures that individuals' scores ``mean the same thing" regardless of the racial group to which they belong, then the right reference class \emph{for every individual} must be the group defined by their racial status. The presumption here is that race gives the {finest-grained} statistical grouping that contains all and only relevant information about every single scored individual. That is, it's not just that the reference class for Jamal is the grouping ``Blacks" but that for Emily, it is ``whites." If this is so, then {no} additional information presented about \emph{any} individual—be it news that Jamal is over 65, or the fact that Emily is sexed female and has a disability—should lead a rational agent to update her belief about the individual's probability. 

This framing of the reference class problem, which focuses on the evidential value of additional information, is best understood in terms of conditional probabilities. Which conditional probability based on an individual's group membership has the best claim to being the probability for that individual? Putting it this way brings out a connection to calibration's definitions. That \emph{calibration within groups} expresses a conditional probability is easily brought to mind: the probability of some outcome ($Y$) for an individual ($X$) given the fact that they have a certain score ($r$) and belong in a certain group ($d$): $P(Y = 1 \mid s(X) = r, X \in d)$. Easier to forget is the fact that classic calibration refers to a conditional probability, too, only one that lacks any features-based group membership information: the probability of an individual's outcome given the fact that the algorithm has assigned them a certain score: $P(Y = 1 \mid s(X) = r)$. 

Comparing the two conditional probabilities shows that calibration within groups simply provides information about probabilities for narrower groups than does classic calibration—conditionalizing not just on a score-based group $r$ but also a features-based group $d$. This suggests that if classic calibration is weakly informative about the evidential value of a given individual's score—a concern, which raises the specter of systematic differential mistreatment and is often used to motivate calibration within groups—then calibration within groups is unlikely to fare much better just by supplying certain further conditionalizations. Only if these extra group conditionalizations cover the set of all reference classes for the entire population of individuals being scored can calibration within groups speak to each individual's probability of the outcome, ensure their equality, and thereby guard against the unfairness that the Same Meaning picture claims to ward off. But this is a dazzlingly tall order, especially since different individuals likely have different right reference classes. As the number of individuals that an algorithm scores multiplies, so does the number of reference class problems and in turn, the number of groups within which scores must be calibrated for the Same Meaning picture to hold true. That an algorithm might just satisfy calibration within the right set of groups is, to be sure, not impossible, though the \emph{presumption} that it does is highly optimistic, bordering on wishful thinking.

\red{This is not to say that the reference class problem is so daunting a problem that efforts to solve it have been all but abandoned. Proposals toward a solution indeed continue to be developed.\footnote{\red{For recent discussion of such proposals, see} \autocite{wallmann2017four}.} That said, H\'{a}jek is, for his part, broadly pessimistic about \red{such solutions in his landmark paper. As he puts it, while we find ``the reference class problem bobbing up in important versions of every major interpretation of probability... they each need to be supplemented with a further theory about what are the ‘right’ reference classes on which probability statements should be based. Yet I believe that the prospects for such theories (e.g., in terms of ‘narrowest classes for which reliable statistics can be compiled’, or ‘total evidence’) are dim."\autocite[580]{hajek2007reference} Regardless of whether one shares H\'{a}jek's pessimism, what matters for the purposes of this paper is that solving the reference class problem is clearly a formidable challenge. So a motivation for calibration that depends on having the correct answer to it is one that comes with a major caveat, which considerably undercuts its persuasive power.}}

\red{So, the Same Meaning picture presumes a solution to the reference class problem, a notoriously hard problem, a solution that it does not defend and is, as an independent matter, unlikely to be correct. But even setting all this aside, there is a further issue. Calibration's proponent cannot overcome these foregoing problems by helping themselves to a method for solving the reference class problem. To see why, suppose that one \emph{did} find a way to determine the right reference classes for individuals. Suppose, for instance, that one managed to make good on Carnap's recommendation to use one's total evidence and successfully operationalized the idea to arrive at a solution to the reference class problem.\autocite{carnap1947application} Would this salvage the Same Meaning motivation for calibration? It unfortunately would not, because that motivation only arises when one \emph{does not} have the means to solve the reference class problem and so \emph{does not} know what risk scores mean for individual probabilities. Recall what motivates the Same Meaning picture: we are worried about our ability to properly interpret algorithmically-output risk scores. We worry that scores might systematically ``mean different things" for individuals of different groups, and so we worry that we might treat individuals differently by misinterpreting what their individual risk scores mean for their true individual probability. But we would not have such worries in the first place if we could derive a solution to the reference class problem. For then we would simply know what individual scores ``mean" because we would know what reference classes to use to determine their individual probabilities. But then the Same Meaning story about calibration's normative significance would ring entirely hollow.\footnote{\red{I thank an anonymous reviewer at \emph{Philosophical Studies} for helping me to see this point.}}}

\red{This shows that the Same Meaning story for calibration's significance as a fairness criterion only works in a rather narrow band of epistemic circumstances. On the one hand, one must know that the risk scores output by an algorithm are calibrated within a set of groups which include the right reference classes for the individuals that one comes to face with. So, one must have an \emph{answer} to the reference class problem for these individuals in order to know that calibration within those groups indeed ensures that same scores ``mean the same thing." But on the other hand, one must \emph{not} have a \emph{general method} for solving the reference class problem that could be operationalized in this setting. For if one could derive the right reference class for each individual and simply look up, as it were, the group probability for that reference class, then one would not need calibration within groups to guard against inconsistent score interpretation. One would be able to directly determine the correct probability for each individual, and the assurance offered by calibration would be extraneous.}

A reply might now appear, one that objects to my characterization of calibration’s connection to fairness. Calibration, its defender claims, guards against a type of unfairness related to \emph{groups}, not individuals. I will now address two variants of this response. The first clarifies that the Same Meaning claim applies to the meaning of group not individual scores (on average): that calibration within racial groups, for instance, ensures that scores for whites taken as a group “mean the same thing” as scores for Blacks as a group. The second modifies the notion of fairness that calibration targets. Calibration within groups, this reply claims, ensures that the risk tool is not unfair to individuals \emph{in virtue of their membership in a particular group}. 

\section{A different kind of fairness} 

One way to escape the reference class fallacy charge pressed in the preceding section is to deny that the Same Meaning picture of calibration relies on an extrapolation of group statistics to individual probabilities. Perhaps calibration’s normative significance persists even as it speaks only to group probabilities or group-based assurances of fairness. 

\subsection{The Same Meaning picture for \emph{groups}}

Indeed, a particular individual’s score of 8 does not “mean the same thing” on average as another individual’s score of 8, simply because the two individuals belong to groups within which scores are calibrated. To conclude so, this defender of calibration concedes, is to commit a reference class error. Fortunately, she says, the Same Meaning picture claims only that calibrated scores \emph{for those groups} mean the same thing, e.g., that the group of individuals who are women and those who are men on average have scores that mean the same thing. Or that two individuals, drawn randomly from the pool of women and men, with the same score have in expectation the same probability of the outcome. 

First, it bears noting that most “meaning”-based talk of calibration that appears throughout the literature does not voice the Same Meaning picture in terms of what “group scores” mean or what the score of the “average” member of a group means. I suspect the reason that scholars have avoided expressing the idea in this way is not just because doing so would be cumbersome. Rather, it is not clear what “group scores” are or who the “average” member of a group is. These quantities may be defined mathematically but do not refer to any actually existing scores or persons and so seem to bear a tenuous connection to matters of fairness. 

Take the claim that women as a group have scores that mean the same thing as the scores of men as a group. It is hard to pin down what such a claim means, because there are no such things as “woman scores.” The algorithm does not assign scores to \emph{women}. No doubt that there are scores that individuals who are women have. And certainly their scores may be pooled together to generate a population that may be characterized statistically, and statements can be made about these scores, including statements such as calibration. But these individuals do not have their scores “as women," and so \red{these statistical conditions should not be translated into claims that suggest there is a further kind of score---say, a special ``\emph{qua} woman score"---that exists above and beyond the individual scores that comprise the group.}\footnote{``Group score" thinking is suggested by statements such as the following by Kleinberg et al. They write that the calibration condition ``asks that scores mean what they claim to mean, even when considered separately in each group... [L]ack of calibration within groups on bin b means that these people have different aggregate probabilities of belonging to the positive class." But what exactly is it to consider a score ``in each group"? The proposal that we consider individuals \emph{qua} members of some group edges toward interpreting particular scores as instances of mysterious ``group scores." As puzzling is the claim that ``people have different aggregate probabilities of belonging to the positive class." I take this to refer to an aggregation across different individuals in the group, but in what sense does such an ``aggregate" quantity belong to a particular individual? There, again, appears a slippage from some fact about the group to some fact about the individual. \autocite[4]{kleinberg2016inherent}.}
  
The revised group-based Same Meaning picture also expresses a different kind of guarantee about fairness, one that seems to me to have considerably weaker normative force. In granting that calibration cannot speak to the meaning of particular scores, it abandons the view of calibration as offering guidance to the acts of interpreting, comparing, and making decisions on the basis of actual individuals’ scores.\footnote{It does offer guidance for reasoning about “average” members of a group, but I am skeptical of the normative significance of this. I return to this matter in \S 5.} So, equalizing the meaning of women’s compared to men’s scores or the average woman’s compared to the average man’s score cannot guard against the kind of disparate mistreatment of individuals that exponents of calibration typically claim. Meanwhile, it is hard to see the normative appeal of the group-based assurance that calibration \emph{can} make. It is cold comfort to individuals being scored by an algorithmic risk tool that per its satisfaction of calibration within groups, members of one of the groups to which they belong are on average scored similarly to members of another group. Or that the “average” member of that group with their score has the same probability of the outcome as the “average” member of another group with the same score. 

Yet at times, it seems that proponents of the Same Meaning view take there to be normative value in precisely such a guarantee: that calibration guards an individual from a certain kind of unfairness, pertaining to their membership in \emph{this} group. I now turn to show why this claim too cannot hold. 

\subsection{Fairness in virtue of \emph{group membership}}

Might calibration within groups bear on something weaker than fairness to individuals but stronger than a mere group-based guarantee? Indeed, Hedden’s paper explicitly sets out to target this Goldilocks kind of fairness, which he compares favorably to fairness to individuals (taken to be too broad for statistical criteria to be able to speak to) and fairness to mere groups (taken to be too normatively weak). About fairness “in virtue of [] group membership,” he writes:
\begin{quote}
How does this notion of fairness differ from the others? One can be unfair to an individual without being unfair to them in virtue of their group membership, for instance if one treats them worse for no reason at all or for some reason unrelated to their group membership, such as their poor eye contact during an interview (assuming that this is in fact unrelated to their membership in various demographic groups). As for unfairness to groups, it is not obvious that fairness is owed to groups, as opposed to individuals… one can perhaps be unfair to an individual in virtue of their membership in a certain group without being unfair to that group itself, for instance if one treats a single individual worse because of their race or gender but at the same time takes other actions that are to the net benefit of that group.\autocite[213]{hedden2021statistical}
\end{quote}
Hedden’s defense of calibration, or at least his unwillingness to rule it out, as a necessary criterion for fairness refers to this notion of fairness in virtue of groups. To quote him again invoking the Same Meaning picture, if calibration were not satisfied, “[a] given risk score, intended to be interpreted probabilistically, would in fact correspond to a different probability of being positive, depending on the individual’s group membership. This seems to amount to treating individuals differently in virtue of their differing group membership.”\footnote{Ibid., 225--226.}

This discussion of calibration and its connection to ``fairness in virtue of groups" suffers two defects. There is first the reference class problem already discussed. To assume that a given score “corresponds to a different probability of being positive, depending on the individual’s group membership” is to assume that the score’s meaning derives from probabilistic facts pertaining to \emph{that} group. This statement expresses an antecedent commitment to a particular reference class without argument and so is unjustified. Furthermore, we have reason to be highly skeptical that the groups within which scores are calibrated cover the set of all reference classes. 

But we might now see another, an even deeper, way in which Hedden’s argument is flawed. \emph{Inscribed within} Hedden's favored sense of fairness for the question of whether there might be any statistical criteria of fairness is a kind of reference class fallacy. To see what is problematic in the notion of “fairness in virtue of group membership,” consider what it is to treat individuals in virtue of their group membership. If I make a decision about you based on your score alone, in what way do I treat you, fairly or unfairly, \emph{in virtue of} some group to which you belong? Assuming that I am not reasoning about your group membership in interpreting your score, what would it be to treat you in this context \emph{qua} Black person as opposed to \emph{qua} Black female person or \emph{qua} 35 year-old person? \red{After all, I am acting just on the basis of your score, so how exactly does my treatment bear this connection to any particular group to which you belong?}

Thus, even if we grant that the reference class problem is resolved and an individual's score is indeed wrong in the way suggested by miscalibration within some group, there is a further error in Hedden’s reasoning about what faulty treatment in these circumstances would mean. His claim is that such differential treatment would be a case of treating an individual differently (unfairly) \emph{in virtue of} their membership in that group. The implicit assumption here seems to be that the reason why the individual’s score reflects an erroneous probability is because they belong to that group. The thought goes something like this: If an individual is treated according to the probability that is suggested by their biased score, and their score is biased because they belong to some group, then their unfair treatment will be because of their group membership and so they will have been treated unfairly in virtue of their group membership. 

But what is it for an individual’s algorithmically generated score to be biased \emph{because of their group membership}? Again, it is not as though the algorithm approaches an individual for whom it must make some prediction \emph{as a woman} or \emph{as white}. In fact, predictive algorithms earn their keep precisely by building complex models that make use of an individual's many features---not by first clocking them as members of a given group and then treating them accordingly, as the notion of ``treatment in virtue of group membership" suggests. There are cases, to be sure, where algorithmic systems show poorer predictive performance on certain groups, because they are numerically few and outcomes for such groups co-vary with features in ways different than the dominant pattern captured by the model. In these cases, we may draw on a broader set of social and technical facts to conclude that the reason why the system fails to perform as well for an individual is \emph{because} they belong to that minority group. And there are of course also those cases where algorithms explicitly use group membership to determine the scores they output. \red{But these are all cases where algorithmic treatment is “because of” group membership in ways that are orthogonal to satisfying or failing to satisfy the statistical criteria at the center of discussions of algorithmic fairness.\footnote{As Hedden notes at 214 in ``On statistical criteria of algorithmic fairness."} No purely statistical criterion for fairness may pronounce that an algorithm is fair or unfair to individuals in virtue of their group membership, because no statistical criterion may reveal whether an instance of computing and acting on an individual’s score is an instance of computing and acting on their score \emph{qua} member of some group. To think that one could reason from a particular instance of unfair score interpretation to the group ``in virtue of which" the individual was treated unfairly is to make an error rather similar to the reference class fallacy: to think that a particular case of unfair individual treatment is an instance of a particular kind of unfair group treatment, simply because the individual may be classified as a member of that group.}\footnote{Long appears to commit the same error. Discussing a case in which a threshold decision policy is used with calibrated scores, generating lack of group parity in false positives, he writes, “To be sure, these individuals can complain that an error has in fact been made. But they cannot argue that the false positive rate shows that they were subjected to a higher risk in virtue of their group membership.” The implication here is that in the case of a violation of calibration (Long's foil to false positive rate) the individual \emph{would} have been subjected to a higher risk in virtue of group membership. ``Fairness in machine learning: Against false positive rate equality as a measure of fairness," 71.} 

\section{Extending the challenge}
\red{I’ve shown that the Same Meaning picture, the predominant normative argument for calibration as a statistical criterion of algorithmic fairness, depends crucially on an inference from group probabilities to individual probabilities. To make this inference properly is to successfully solve the reference class problem. Thus, the Same Meaning picture effectively claims a solution to this daunting problem---a solution that is, as I have argued, flawed in several ways. First, it is a solution that is simply presumed rather than argued for. Second, it is a solution that is very likely incorrect. And third, if the solution were the result of a general method for solving the reference class problem, then this method would obviate the need for the Same Meaning motivation for calibration in the first place.}

It bears saying explicitly at this point that my argument does not serve to undercut calibration's normative significance in toto; the condition might still be justified by some other means, and I leave it to future work to propose other ties that calibration might have to fairness.\footnote{For starters, those sympathetic to group fairness guarantees may be unfazed by these defects in ``meaning"-based talk. Similarly, those who take calibration in the spirit of a diagnostic that may guide efforts to uncover other non-statistically defined types of unfairness are also likely to remain steady in their support of calibration.} In the meantime, if I am right that calibration's distinctive normative advantage over its competitors hangs on something like the Same Meaning picture, my challenge to it should lead us to revise our sense of its privileged status and return to the open field of fairness metrics to think anew about which conditions might best (if not perfectly) achieve our aims.

We are now in a position to cast a wider view onto the problem of statistical criteria of fairness---but our path will make one final detour through calibration, this time an influential objection to it recently pressed by Benjamin Eva, which bears some loose connection to mine.\autocite{eva2022algorithmic} Although Eva also points to the existence of multiple groups to undercut calibration's connection to fairness, he does not appear to recognize the reference class problem. This is evinced later on in his paper, when he commits a similar reference class fallacy when defending his own preferred statistical criterion for fairness. Over the course of walking through Eva's argument and distinguishing mine from it, I will step back to extend my challenge and make some methodological points about the risks and pitfalls of the dominant approach to theorizing statistical criteria of algorithmic fairness based on toy cases.  

\subsection{Eva’s argument}

In an argument on the “non-necessity of calibration,” Benjamin Eva presents a case in which an algorithm predicts risk for drivers, who are either young or old, according to their credit score.\AtNextCitekey{\clearname{author}}\autocite{eva2022algorithmic} The algorithm violates calibration within age groups, because its assignment of a risk score of $\frac{1}{20}$ corresponds to a probability of $\frac{3}{80}$ for young drivers and $\frac{1}{40}$ for old drivers, thereby marking old drivers who are in fact less risky than young drivers ($\frac{1}{40} < \frac{3}{80}$) as equally risky. Still, Eva claims that despite this violation of calibration, the tool is not unfair to old drivers, since for those assigned a score of $\frac{1}{10}$, the algorithm is miscalibrated \emph{in favor of} old drivers. Eva’s example is therefore constructed so that the two cases of miscalibrated scores within age groups offset each other. He concludes that the “algorithm does not systematically treat younger drivers more favorably than older drivers or vice versa” and so is fair, even while breaking with calibration within age groups.\autocite[249]{eva2022algorithmic} 

Eva uses the case to raise the following questions for the defender of calibration: For which groups is calibration a requirement for fairness? Is the fairness-relevant grouping \emph{old drivers} versus \emph{young drivers}? Or is it \emph{old drivers with good credit scores} versus \emph{young drivers with good credit scores}? Need a tool be calibrated across \emph{all} groupings for it to be fair? If not, which ones matter? Eva’s own view is that requiring some statistical criterion to hold for all groups is too demanding to be a genuine requirement for algorithmic fairness and that it is more reasonable to, instead, determine from the outset which groups are “significant” for matters of fairness and ensure that these groups are not subject to unfair treatment.\autocite{eva2022algorithmic} So we might rule that because of “important social, political, economic and historical origins and ramifications,” age-defined groups matter but not age and credit scores-defined groups.\autocite[251]{eva2022algorithmic} To pick out the fairness-relevant groups is to choose the fault lines across which treatment must not systematically differ and to allow differential treatment to track finer distinctions within those groups so long as the differences offset one another upon zooming out to consider the coarser group as a whole. 

To briefly return to a matter discussed earlier, we might wish to pause to ask what appeal this picture of group fairness has. It seems to me rather undesirable that Eva’s notion of fairness allows systematic differential treatment along lines that are not antecedently marked out as “significant” so long as they wash out in the broader ``significant” group. Consider how little consolation lies in the assurance given to the old driver with a good credit score who—as Eva himself admits—is treated unfairly relative to his more youthful counterpart that, really the algorithm is fair, since old drivers with a bad credit score are undeservedly treated \emph{better} than their younger counterparts. This reassurance falls flat because, as I suggested in \S 3, the notions of fairness that Eva targets here, fairness \emph{to groups} and fairness \emph{overall}, in failing to engage second-personal reasons and respect differences among persons, lack normative force---at least as compared with the Same Meaning account of why calibration matters for fairness.\footnote{Hedden also worries about the normative significance of purely group-based notions of fairness and articulates a similar concern about their allowing different instances of unfair treatment to “offset” each other. ``On statistical criteria of algorithmic fairness, 213.}

\red{Eva's counterexample works by adding an additional group (age groups \emph{and} credit score groups) to complicate what it takes for risk scores to satisfy calibration within groups. Thus, it already contains most of the resources needed to attack the stronger case for calibration expressed in the Same Meaning picture.} He need only modify the case slightly and reframe it in terms of reference classes. That version of the objection (whose details I defer to the footnote) puts forth a set of scores that are calibrated within age groups but miscalibrated within age and credit score groups.\footnote{Assuming, as Eva does, that the four age and credit score defined groups are equally sized, the algorithm represented below satisfies calibrated within age groups (the base rate for young drivers with a score of 1/10 is 3/80; the same is true of old drivers) but violates calibration within age and credit score groups (the score of 1/10 corresponds to 1/40 for young drivers with good credit scores, 1/20 for young drivers with bad credit scores, 3/80 for old drivers with bad credit scores).
  \begin{center}
  \begin{tabular}{ l l c || c }
  Age & Credit & Risk Score & Base Rate  \\ \midrule
  Young & Good & $\frac{1}{10}$ &  $\frac{1}{40}$  \\ \noalign{\smallskip}
  Young & Bad & $\frac{1}{10}$ & $\frac{1}{20}$ \\ \noalign{\smallskip}
  Old & Good & $\frac{1}{20}$ & $\frac{1}{80}$ \\ \noalign{\smallskip}
    Old & Bad  & $\frac{1}{10}$ & $\frac{3}{80}$\\ \noalign{\smallskip}
\end{tabular}\end{center}} This raises the question of what we should say about the “meaning” of the risk scores of two drivers, one old with a bad credit score, the other young with a bad credit score. If they may be classified according to groups that are calibrated and groups that are miscalibrated, do their identical scores “mean the same thing” or “mean different things”? This brings out the reference class problem.

One reason for why Eva does not go this way is that he does not target fairness to individuals or fairness to individuals “in virtue of” their group membership. By restricting his discussion to fairness to groups, he circumvents any need to discuss individuals, without which the reference class problem does not appear. But that is not, I think, the only reason. I want to now show that Eva’s own preferred statistical criterion, which he calls “base rate tracking,” trades on similar intuitions about fairness that owe their strength to something like the Same Meaning picture and so, similarly, illicitly helps itself to a solution to the reference class problem. He writes:
\begin{quote}
[B]ase rate tracking is motivated by a natural philosophical intuition regarding the nature of fairness that any difference in the way that an algorithm treats two groups needs to be justified by a corresponding difference in the relevant behaviors/properties of the two groups. It is unfair to treat white loan applicants as if they have a much lower average risk of defaulting compared to black applicants if they do not actually have a much lower default rate.\autocite[259]{eva2022algorithmic}
\end{quote}

This passage, like much “meaning”-based talk of calibration, makes reference to a set of scored individuals—in this case, white loan applicants and black loan applicants—in a way that is ambiguous between discussing them as a subpopulation defined by that feature or as a pluralization of individual persons, each of whom have some feature and some risk score. I read Eva (and others in the literature who make use of similar locutions) in the vein of the latter, since algorithms strictly speaking “treat” individuals not groups, and it's not clear what abstract group scores \emph{are} let alone mean. Scored individuals may be pulled together into different populations based on their features, and sometimes, we may want to make claims about the distribution of scores assigned to the individuals who comprise that subpopulation, saying as Eva does in the passage above, for instance, that the algorithm “treat[s] white loan applicants as if they have a much lower average risk of defaulting compared to black applicants.”\autocite{eva2022algorithmic} Nevertheless, one should bear in mind that this is ``group" treatment only in a loose sense. It is properly understood as treatment of a collection of individuals who all share a particular feature, that of being racialized “white” or “black.”

If we read Eva in this vein, as referring to individuals who have a certain feature, the preceding statement may be reformulated without loss of fidelity as, “It is unfair to treat a white loan applicant as if they have a much lower average risk of defaulting compared to a black applicant if they do not actually have a much lower default rate.” This paraphrase makes clear that it is individuals who receive treatment and in so doing, brings out how Eva’s motivation for base rate tracking closely resembles “meaning”-based motivations for calibration. It is unfair, Eva claims, to treat two individuals who in fact have similar default rates as though the white applicant has a much lower risk of default than the black applicant. In other words, it is unfair to assign the white applicant and black applicant two different scores, if they in fact have similar probabilities of default. To put the idea in terms of scores' ``meaning": to do so would risk a score-interpreter's taking the two different scores to “mean different things” when they actually “mean similar things.”  

I share with Eva the intuition that this treatment would be unfair. The problem, as I have argued throughout this paper, is that this conclusion about how individuals should be treated is based on an inference about what their score “means” or what their “true risk” is, which is derived from a set of group statistics, \red{i.e., base rates. But these group base rates may not apply to an individual without a prior determination that \emph{that} group is the relevant reference class for them. The claim that the individuals have similar probabilities of default thus assumes in this example that \emph{racial} base rates are the right reference class for determining individuals' probabilities of loan default. That presupposition is made but not defended to pump the intuition for base rate tracking’s connection to fairness, just as we saw was the case for the Same Meaning motivation for calibration.} Eva’s base rate tracking criterion thus either draws its normative force from a view that suffers the same reference class problem as does calibration’s Same Meaning picture, or it speaks only to a group-based notion of fairness, which for one, has considerably weaker normative significance, and for another, is a notion of fairness different than the one Hedden targets and for which he defends calibration as a possibly necessary requirement.

\subsection{Methodological lessons}

Eva’s discussion of calibration and base rate tracking presents a natural entry point into a wider methodological conversation about the dominant approach that philosophers have thus far taken to \red{theorizing what statistical criteria are necessary and/or sufficient for an algorithm to be fair.} The standard approach in the literature proceeds as follows: An example is drawn up of a toy algorithm that intuition judges to be fair or unfair. It is revealed that the algorithm does or does not satisfy some statistical criteria. So, the criteria are ruled out as not necessary or not sufficient for fairness.

There are a number of risks and pitfalls to focusing on highly stylized cases as guides for reasoning about fairness in algorithms. \red{Before moving on to discuss them, I want to first say explicitly that I do not mean this to be a wholesale criticism of this methodology. Most of the argument of this very paper takes a similar approach to issues in algorithmic fairness; and to the extent that it manages to make progress on some of these matters, any positive contribution will be in part due to these methods. Moreover, I take those works that follow these methods which I discuss throughout this paper to have significantly advanced our understanding of algorithmic fairness. So, it is self-evident that the methods have been very fruitful indeed. Still of course, no methods are perfect and foolproof. The following methodological reflections emerge out of consideration of the sources of some of the problems discussed in the previous sections. Thus, my aim in this section is to probe what aspects of the methods may have encouraged these pitfalls, so that we may learn from them and, hopefully, be better able to avoid them in the future.}

First, as I earlier mentioned, toy examples in the literature often assign scores to individuals who are represented as having only one or two features. In these cases, the reference class problem does not present itself, because it does not exist at all, as the fictionalized individuals may only be classified according to a single group. Real cases of algorithmic risk scoring, by contrast, do not involve individuals so thinly characterized, and there, the reference class problem is a genuine problem. \red{Thus, intuitions and analyses forged under overly simplified cases may range from being misleading to fallacious when used to draw conclusions about what the satisfaction of group statistical criteria means for the fairness of some algorithm in the real world.\footnote{\red{I take Eva's challenge to calibration discussed earlier to be an instance of this methodological lesson in action. Eva proceeds by putting forth a toy example of an algorithm that uses two rather than just one feature and showing how this slightly more complex case presents problems for calibration that do not emerge in single-feature examples. \red{This argument thus illustrates both the limits and value of toy examples in this literature.}}}}

Second, it bears keeping in mind that toy cases are well-crafted dossiers of fact, which make certain facts highly salient while others are excluded. Thus, their construction and presentation strongly shape our intuitions about fairness. For an illustration of how, return to Eva’s counterexample to calibration. I’ve reproduced the table of this case from Eva’s paper below in Fig. 1. 

\begin{figure}[h]
\centering
\includegraphics[scale=0.8]{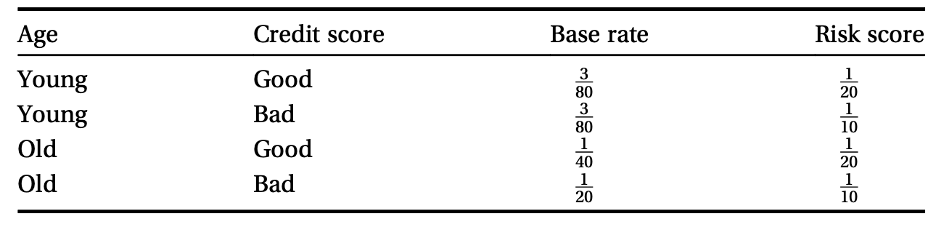}
\caption{Eva's table from ``Algorithmic fairness and base rate tracking," 249.}
\end{figure}
Eva's algorithm makes risk assignments according to individuals’ credit scores alone: $\frac{1}{20}$ if the driver has a good credit score; $\frac{1}{10}$ if they have a bad credit score. Recall that he takes this algorithm to be fair to young and old drivers alike as it does not systematically score either group taken as a whole more favorably than the other, and he expects the reader to share this intuition. But I am skeptical of that intuition, both whether it is as widely shared as Eva presumes and whether even upon granting it, our approval of the algorithm in this highly abstracted case would track our considered judgments about fairness in real cases where an algorithm assigns scores in this way. To begin with, one might find fault with the fact that the algorithm is geared only to detecting differential outcomes among old drivers. Its predictions are based on a criterion that does not track any difference in accident rates among young drivers. The choice to go with a predictor that draws a line among young drivers with no statistical basis seems to me unfair to them indeed, and moreover, it is an unfairness that is identifiable via statistical facts alone. How outcomes among young drivers vary falls to the wayside entirely in favor of a criterion that “works” for old drivers.\footnote{Indeed, this is a concern about the algorithm's violation of calibration within age groups.}

This critique of the algorithm—the fact that it bases its assignments of risk on a factor that is differentially good for young and old drivers—is easily taken for granted in stylized cases like these, because the algorithm and individual/group data are presented in a highly decontextualized form. The table above encourages us to conceive individuals as only having two features, an age and a credit score. It is unsurprising that under circumstances of such limited information, basing risk determinations on credit scores alone might not seem so unfair. After all, we might be led to shrug off the concern and say: what else does this algorithm have to go off of? Of course, in real life, algorithms are powerful precisely because they have a lot to go off of. They have access to many other data and may model outcomes in complex ways to fit patterns that are predictive for young and old drivers alike. I suspect that it is because we think that algorithms typically \emph{can} find correlations that “work” for drivers of all ages that we might judge an algorithm that assesses risk on the basis of a feature that works only for old drivers to be unfair to young ones. This fact, however, is easily lost when considering abstracted toy cases. Our intuitive judgment that the algorithm is fair might therefore trace to the fact that the algorithm makes straightforward use of the little information it appears to have and its doing so does not disadvantage each group as a whole. But were an algorithm to generate these risk scores in real life, we might deem it unfair to different age groups in its complete neglect of any structure in the pattern of accident rates among young drivers.\footnote{Even the presentation of “base rate” data in this table risks the misleading because unwarranted thought that credit scores \emph{explain why} old drivers with bad credit scores as a group have worse driving outcomes than old drivers with good credit scores. Base rates are simply group-based probabilities, and when there is outcome variation within some population, partitioning that population along any feature will likely generate different “base rates.”}

\red{Finally, a point that does not trace back to the use of stylized cases in particular but applies more broadly to this strand of work.} There is a persistent lack of clarity about the kind of intuition about “fairness” that is meant to be drawn out in these cases. In his section on the non-necessity of calibration alone, Eva makes reference to several distinct notions: being fair to young and old drivers, fair to young drivers with good credit scores, fair “overall,” being unjust.\footnote{\autocite{eva2022algorithmic}, \S III.A.} And there are many potential targets besides. When our intuitions are elicited about the fairness of toy cases, it is not clear which of them (if any) our intuitions are speaking to. Indeed, different authors seem to prefer different notions. Eva considers fairness to groups; Hedden, fairness to individuals in virtue of group membership; Long, procedural fairness; Loi and Heitz, non-discrimination; meanwhile, Kleinberg et al. and Corbett-Davies et al., are typical of contributors to the technical literature in keeping it crisp at just “fairness.” Nevertheless, despite such variation, scholars proceed as though they are interlocutors all speaking on the same topic. They respond to each other’s cases and arguments in rebuttal, in whole-hearted support, in a similar spirit if not total endorsement—even though it often remains unclear when these objections, rejoinders, and gestures of support in fact make contact. 

To reiterate, all this is not meant to be taken as a call for the end of models in discussions of algorithmic fairness. Rather, my concern is to recall the simple yet always pertinent methodological insight from the philosophy of science about good modeling practice: that the value of a model is so much a matter of its being at the right level of abstraction. \red{It behooves us to be sensitive to when the models we use to draw conclusions about fairness might be idealizing away too many of the epistemically and normatively significant features of the target of our theorizing, which remains, as ever, real predictive algorithms used on real people.} 


\section{Conclusion}
It is the mark of a good philosophical picture that it is suggestive in the right way: that it draws out what is significant, encourages fruitful inferences, and situates an idea within a context that illuminates its value. The Same Meaning picture of calibration does this by transporting the reader to the stage of score \emph{interpretation} and calling to mind the essential task of \emph{comparison}. It thereby evokes a vision of algorithmic fairness concerned primarily with procedure and the individuals who face up with that procedure. 

The problem with this picture that I have largely focused on in this paper is that it simply does not hold up as an account of the normative property of fairness that calibration within groups can in fact secure. This is because of the problem of reference classes, which afflicts not just the extent to which calibration but any group-based statistical criterion may speak to fair treatment of individuals (or fair treatment of individuals ``in virtue of" their group membership). 

My guess is that the source of this mistake in reasoning lies in the difficulty inherent to the task of translating mathematics to ordinary language. In particular, it remains a real challenge for anyone writing on algorithmic fairness to speak about formal statistical criteria in an intuitive manner that conveys their connection to key ethical notions while staying in full keeping with their mathematical definition. Locutions such as “drawn randomly from the distribution of the population of women” or “probabilistic independence, conditional on discretized score” are, in addition to being unwieldy, simply difficult to make sense of. Caught between, on the one hand, communicating solely in mathematical notation and fumbling with what can begin to sound like legalistic caveat-filled talk, on the other, no wonder contributors look for a third way. So enter the quick glosses and tidy metaphors that might let slip crucial details through the cracks. I have shown in this paper that the reference class problem is one such detail that has slipped through with little notice. 

But I want to close on a second point of unease with the Same Meaning picture. The Same Meaning picture expresses a particular normative orientation towards score interpretation: that one should interpret an individual’s score’s meaning against the background of their belonging to a particular group. This involves applying a particular formula of abstraction. We are encouraged to see this person as an instance of some group and to pronounce based on calibration facts about that group that they are, in expectation, treated fairly or unfairly qua their belonging to that group. 

This orientation is rationally worrisome (per the reference class problem); as it is also, I think, ethically so. The issue lies in the caricatured structural analysis exemplified in the Same Meaning picture and indeed any picture that utilizes group-based statistical facts to apply a filter to a person, such that I may now see you qua average Black person, now qua average woman, now qua average young person, now qua average young woman; according to which, you are treated unfairly, fairly, fairly, unfairly. “Average” persons are, of course, statistical fictions in the sense that they do not designate any actually existing persons in the real world. Nor is it right to think of an individual as a kind of composite of “averages,” which may be peeled apart, each representing a projection of them into some subspace of identity. To mistakenly think and treat them as much is to assimilate them into a given category and treat them in a manner that some philosophers have argued is an affront to their individuality.\footnote{See e.g., \autocite{eidelson2020respect}.} But one need not go the route of exalting a person’s unique personality and autonomy to take issue with this orientation. I personally am more ill at ease with another ethical pitfall risked by the Same Meaning picture: that it, in encouraging us to see individuals as abstracted members of some group, might assess what fairness demands vis-à-vis membership in that group according to dominant rather than subordinated positions within the group. I take this to be a core insight of intersectionality theory: that analyses which foreground the “average”  or ``representative" person in a group systematically neglect those in the group who are most marginalized.\footnote{The idea here is one of ``intersectional invisibility," which describes the risk of overlooking individuals who belong to multiple marginalized social groups, because they are taken to be unprototypical. See \autocite{sesko2010prototypes}. The founding texts in intersectionality theory are Kimberl{\'e} Crenshaw's \AtNextCitekey{\clearname{author}}\autocite{crenshaw1989demarginalizing} and \AtNextCitekey{\clearname{author}}\autocite{crenshaw1991mapping}.}

And yet it seems to me still true that the statistical fictions of \emph{averages} and \emph{base rates} and \emph{risk distributions} bear on what it takes to treat real-life persons fairly. If this is so, then our task is to figure how we might be able to reconcile these two viewpoints: the abstracted statistical view of a group and the concrete view of an actual person. This is yet another matter of algorithmic fairness on which we are likely to find ourselves pulled in different directions. I can only hope that, at least as compared with the reference class problem and the most infamous and intractable conflict in the literature, the impossibility theorem, our prospects here are higher.\footnote{The impossibility theorem describes a set of results showing that except in marginal cases, an algorithm cannot satisfy multiple popular statistical criteria of fairness at once. First proofs of the theorem appeared in \autocite{chouldechova2017fair,kleinberg2016inherent}.}

\pagebreak

\printbibliography

@article{hedden2021statistical,
  title={On statistical criteria of algorithmic fairness},
  author={Hedden, Brian},
  year={2021},
  journal={Philosophy \& Public Affairs},
  pages = {209--231},
  volume ={49},
  number = {2}
}

@article{hellman2020measuring,
  title={Measuring algorithmic fairness},
  author={Hellman, Deborah},
  journal={Virginia Law Review},
  volume={106},
  number={4},
  pages={811--866},
  year={2020},
  publisher={JSTOR}
}

@article{eva2022algorithmic,
  title={Algorithmic fairness and base rate tracking},
  author={Eva, Benjamin},
  journal={Philosophy \& Public Affairs},
  volume={50},
  number={2},
  pages={239--266},
  year={2022},
  publisher={Wiley Online Library}
}

@inproceedings{loi2022calibration,
  title={Is calibration a fairness requirement? An argument from the point of view of moral philosophy and decision theory},
  author={Loi, Michele and Heitz, Christoph},
  booktitle={Proceedings of the 2022 ACM Conference on Fairness, Accountability, and Transparency},
  pages={2026--2034},
  year={2022}
}

@inproceedings{kleinberg2016inherent,
  title={Inherent trade-offs in the fair determination of risk scores},
  author={Kleinberg, Jon and Mullainathan, Sendhil and Raghavan, Manish},
  booktitle={Proceedings of the 8th Conference on Innovations in Theoretical Computer Science (ITCS)},
  year={2017}
}

@article{mayson2019bias,
  title={Bias in, bias out},
  author={Mayson, Sandra G},
  journal={The Yale Law Journal},
  volume={128},
  number={8},
  pages={2218--2300},
  year={2019},
  publisher={JSTOR}
}

@article{long2021fairness,
  title={Fairness in machine learning: Against false positive rate equality as a measure of fairness},
  author={Long, Robert},
  journal={Journal of Moral Philosophy},
  volume={19},
  number={1},
  pages={49--78},
  year={2021},
  publisher={Brill}
}

@book{simons2023algorithms,
  title={Algorithms for the People: Democracy in the Age of AI},
  author={Simons, Josh},
  year={2023},
  publisher={Princeton University Press}
}

@article{pleiss2017fairness,
  title={On fairness and calibration},
  author={Pleiss, Geoff and Raghavan, Manish and Wu, Felix and Kleinberg, Jon and Weinberger, Kilian Q},
  journal={Advances in neural information processing systems},
  volume={30},
  year={2017}
}

@article{hajek2007reference,
  title={The reference class problem is your problem too},
  author={H{\'a}jek, Alan},
  journal={Synthese},
  volume={156},
  pages={563--585},
  year={2007},
  publisher={Springer}
}

@article{chouldechova2017fair,
  title={Fair prediction with disparate impact: A study of bias in recidivism prediction instruments},
  author={Chouldechova, Alexandra},
  journal={Big data},
  volume={5},
  number={2},
  pages={153--163},
  year={2017}
}

@article{eidelson2020respect,
  title={Respect, Individualism, and Colorblindness},
  author={Eidelson, Benjamin},
  journal={Yale Law Journal},
  volume={129},
  number={6},
  pages={1600--1675},
  year={2020}
}

@article{crenshaw1989demarginalizing,
  title={Demarginalizing the intersection of race and sex: A black feminist critique of antidiscrimination doctrine, feminist theory and antiracist politics},
  author={Crenshaw, Kimberl{\'e}},
  journal={University of Chicago Legal Forum},
  pages={139},
  year={1989},
  publisher={HeinOnline}
}

@article{crenshaw1991mapping,
  title={Mapping the margins: Intersectionality, identity politics, and violence against women of color},
  author={Crenshaw, Kimberl{\'e}},
  journal={Stanford Law Review},
  volume={43},
  pages={1241},
  year={1991},
  publisher={HeinOnline}
}

@article{sesko2010prototypes,
  title={Prototypes of race and gender: The invisibility of Black women},
  author={Sesko, Amanda K and Biernat, Monica},
  journal={Journal of Experimental Social Psychology},
  volume={46},
  number={2},
  pages={356--360},
  year={2010},
  publisher={Elsevier}
}

@article{castro2023correction,
  title={The Fair Chances in Algorithmic Fairness: A Response to Holm},
  author={Castro, Clinton and Loi, Michele},
  journal={Res Publica},
  volume={29},
  number={2},
  pages={331--337},
  year={2023},
  publisher={Springer}
}

@book{reichenbach1971theory,
  title={The theory of probability},
  author={Reichenbach, Hans},
  year={1971},
  publisher={University of California Press}
}

@book{venn1888logic,
  title={The logic of chance: an essay on the foundations and province of the theory of probability, with especial reference to its logical bearings and its application to moral and social science, and to statistics},
  author={Venn, John},
  year={1888},
  publisher={Macmillan}
}

@article{beigang2023reconciling,
  title={Reconciling Algorithmic Fairness Criteria},
  author={Beigang, Fabian},
  journal={Philosophy \& Public Affairs},
  year={2023},
   pages={166--190},
  publisher={Wiley Online Library}
}

@article{grant2023equalized,
  title={Equalized odds is a requirement of algorithmic fairness},
  author={Grant, David Gray},
  journal={Synthese},
  volume={201},
  number={3},
  pages={101},
  year={2023},
  publisher={Springer}
}

@article{holm2023fairness,
  title={The fairness in algorithmic fairness},
  author={Holm, Sune},
  journal={Res Publica},
  volume={29},
  number={2},
  pages={265--281},
  year={2023},
  publisher={Springer}
}

@inbook{salmon1971statistical,
  booktitle={Statistical explanation and statistical relevance},
  title={Statistical explanation},
  author={Salmon, Wesley C},
  year={1971},
  pages={29--87},
  publisher={University of Pittsburgh Press}
}

@article{thorn2017preference,
  title={On the preference for more specific reference classes},
  author={Thorn, Paul D},
  journal={Synthese},
  volume={194},
  number={6},
  pages={2025--2051},
  year={2017},
  publisher={Springer}
}

@inproceedings{holtgen2023richness,
  title={On the Richness of Calibration},
  author={H{\"o}ltgen, Benedikt and Williamson, Robert C},
  booktitle={Proceedings of the 2023 ACM Conference on Fairness, Accountability, and Transparency},
  pages={1124--1138},
  year={2023}
}

@inproceedings{roth2023reconciling,
  title={Reconciling Individual Probability Forecasts},
  author={Roth, Aaron and Tolbert, Alexander and Weinstein, Scott},
  booktitle={Proceedings of the 2023 ACM Conference on Fairness, Accountability, and Transparency},
  pages={101--110},
  year={2023}
}

@inproceedings{wallmann2017four,
  title={Four approaches to the reference class problem},
  author={Wallmann, Christian and Williamson, Jon},
  booktitle={Making it formally explicit: probability, causality and indeterminism},
  pages={61--81},
  year={2017},
  organization={Springer}
}

@article{carnap1947application,
  title={On the application of inductive logic},
  author={Carnap, Rudolf},
  journal={Philosophy and Phenomenological Research},
  volume={8},
  number={1},
  pages={133--148},
  year={1947},
  publisher={JSTOR}
}

\end{document}